%% file: main.tex
\documentclass{article}

\usepackage[letterpaper,margin=3cm,footskip=30pt]{geometry}

\usepackage{amssymb}
\usepackage{amsmath}
\usepackage{latexsym}
\usepackage{url}
\usepackage{xcolor}
\usepackage{booktabs}
\usepackage{xspace}
\usepackage{paralist}
\usepackage{hyperref}
\usepackage[linesnumbered,ruled,vlined]{algorithm2e}

\usepackage{amsfonts}       
\usepackage{nicefrac}       
\usepackage{microtype}      
\usepackage{lipsum}
\usepackage{stfloats}
\usepackage{setspace}
\usepackage{multirow}
\usepackage{float} 
\usepackage{subfigure} 
\usepackage{appendix}
\usepackage{pdflscape}
\usepackage{threeparttable} 
\usepackage{rotating}
\usepackage{lineno} 
\usepackage{marvosym} 
\usepackage{silence}
\usepackage{mathrsfs}

\input{macros}

\title{CKGAN: Training Generative Adversarial Networks Using Characteristic Kernel Integral Probability Metrics}

\date{}

\begin{document}

\input{authors}

\maketitle

\input{abstract}
\input{intro}
\input{related}
\input{method}
\input{relation}
\input{exp}
\input{concl}

\input{ack}

\bibliographystyle{abbrv}
\bibliography{refs}

\clearpage

\appendix
\input{appendix}

\end{document}

%% file: macros.tex
\newcommand{\myparagraph}[1]{\noindent\textbf{#1}.\,}

\newcommand{\confversion}[1]{\xspace}

\DeclareMathOperator{\argmax}{arg\,max}
\DeclareMathOperator{\softmax}{softmax}

\newcommand{\set}[1]{\{\, #1 \,\}}

\newcommand{\ckgan}{CKGAN\xspace}
\newcommand{\ckipm}{CKIPM\xspace}
\newcommand{\mmdgan}{MMD GAN\xspace}
\newcommand{\immdgan}{IMMD GAN\xspace}
\newcommand{\smmdgan}{SMMD GAN\xspace}

\newcommand{\veegan}{VEEGAN\xspace}
\newcommand{\gmmn}{GMMN\xspace}
\newcommand{\unrolledgan}{Unrolled GAN\xspace}
\newcommand{\wgangp}{WGAN-GP\xspace}

\SetFuncSty{text}
\SetArgSty{text}
\SetKw{OR}{or}
\SetKw{AND}{and}
\SetKw{NOT}{not}
\SetKw{TRUE}{true}
\SetKw{FALSE}{false}
\SetKw{NULL}{nil}
\SetKw{Break}{break}
\SetKw{Continue}{continue}
\SetKwInOut{Input}{Input}
\SetKwInOut{Output}{Output}
\newcommand{\State}[1]{#1\;}

\makeatletter

\renewcommand{\thesubfigure}{\alph{subfigure}}
\renewcommand{\@thesubfigure}{(\thesubfigure)\hskip\subfiglabelskip}
 
\newcommand{\equationref}[1]{Equation~\ref{#1}}

%% file: authors.tex
\newcommand{\osaka}{$^{1}$}
\newcommand{\osakanagoya}{$^{1,4}$}
\newcommand{\szu}{$^{2}$}
\newcommand{\sics}{$^{3}$}
\newcommand{\sicsszu}{$^{2,3}$}
\newcommand{\nagoya}{$^{4}$}

\author{\osaka Kuntian Zhang\thanks{Work done when studying at Osaka University.}, \szu Simin Yu, \sicsszu Yaoshu Wang, \\ \osaka Onizuka Makoto, \osakanagoya Chuan Xiao\\
\small{\osaka Osaka University, \szu Shenzhen University,}\\ 
\small{\sics Shenzhen Institute of Computing Sciences,}\\
\small{\nagoya Nagoya University}\\
\small{kuntianzhang@gmail.com, 1900271062@email.szu.edu.cn, yaoshuw@sics.ac.cn,}\\ 
\small{\{onizuka, chuanx\}@ist.osaka-u.ac.jp}
}

%% file: abstract.tex
\begin{abstract}


In this paper, we propose \ckgan, a novel generative adversarial network (GAN) variant 
based on an integral probability metrics framework with characteristic kernel (CKIPM). 
CKIPM, as a distance between two probability distributions, is designed to optimize the 
lowerbound of the maximum mean discrepancy (MMD) in a reproducing kernel Hilbert space, 
and thus can be used to train GANs. \ckgan mitigates the notorious problem of mode 
collapse by mapping the generated images back to random noise. To save the effort of 
selecting the kernel function manually, we propose a soft selection method to 
automatically learn a characteristic kernel function. The experimental evaluation 
conducted on a set of synthetic and real image benchmarks (MNIST, CelebA, etc.) 
demonstrates that \ckgan generally outperforms other MMD-based GANs. The results also 
show that at the cost of moderately more training time, the automatically selected 
kernel function delivers very close performance to the best of manually fine-tuned one 
on real image benchmarks and is able to improve the performances of other MMD-based GANs. 

Source codes are available at \url{https://github.com/chuanxiao1983/CKGAN/} .


\end{abstract}

%% file: intro.tex
\section{Introduction}
\label{sec:intro}
In recent years, generative adversarial network (GAN) \cite{goodfellow2014generative}, as one of generative models, has shown superb capability to learn high-dimensional probability distribution. 
It has achieved success in various image-related tasks, such as high-resolution image generation \cite{brock2018large} and image style transfer \cite{chen2020reusing}. 
Whereas diffusion models~\cite{ho2020denoising,karras2022elucidating} have become more popular for these tasks in recent years, it has been shown that GANs still compare favorably against state-of-the-art diffusion models~\cite{huang2024gan}. 

The vanilla GAN \cite{goodfellow2014generative} consists of a generator and a discriminator (also called critic) which are alternately trained in each iteration. The generator transforms a base low-dimensional distribution (e.g. Gaussian distribution) to a generated distribution that approximates the real data distribution. The discriminator distinguishes the generated distribution from the real data distribution. 
The loss function of the discriminator measures the distance between the two distributions, which is typically an $f$-divergence \cite{nowozin2016f} such as Jensen-Shannon divergence for the vanilla GAN, integral probability metrics (IPM) such as Wasserstein distance for WGAN \cite{arjovsky2017wasserstein,gulrajani2017improved}, and maximum mean discrepancy (MMD) for \mmdgan \cite{li2017mmd}.

Many variants have been proposed to improve GAN's performance, especially for those aiming at addressing the mode collapse problem in which the generator fails to produce results with diversity. 
For example: 
\unrolledgan~\cite{metz2016unrolled} introduce a surrogate objective function to imitate a discriminator response to generator changes; 
\veegan \cite{srivastava2017veegan} jointly trains the generator and a reconstructor network that maps the data to noise; 
PacGAN \cite{lin2018pacgan} packs a set of samples as the input data of discriminator.  

In this paper, we introduce characteristic kernel integral probability metrics (\ckipm) based on the IPM framework for learning GANs. 
We define \ckipm as a distance between two probability distributions, and show theoretically that \ckipm is equal to a lower bound of MMD using the characteristic kernel~\cite{fukumizu2008kernel}. 
In contrast to \mmdgan that only considers using a mixture of Gaussian (RBF) kernel, we study the case of integrating a wide range of characteristic kernels applied to the objective function of \mmdgan variants. 
Our contributions are summarized as follows:

\begin{itemize}

  \item We propose \ckgan, a general GAN variant featuring characteristic kernel-based techniques and an IPM framework. The objective to be optimized is defined by a \ckipm, which is a lower bound of MMD and thus can be used to train GANs. In addition, \ckgan alleviates the mode collapse problem by mapping the generated data back to random noise. 

  \item We propose a soft selection method to learn a characteristic kernel, which is a linear combination of multiple characteristic kernel functions whose weights are learned automatically, so that the GAN can generate decent samples without needing human effort that manually picks the kernel function.
  
  \item We conduct experiments on image generation tasks using synthetic and real image datasets (MNIST, CIFAR-10, CelebA, and LSUN church outdoor). We equip \ckgan with six manually tuned kernel options and the automatically selected kernel. The comparisons demonstrate that \ckgan generally outperforms other GANs (\veegan, \mmdgan, 
  \smmdgan~\cite{Arbel2018OnGR}, and \immdgan~\cite{wang2018improving}). The results also demonstrate that the automatically selected kernel performs almost as well as the best of manually fine-tuned ones on real image datasets while consuming moderately more training time, and it is able to improve the performance of various MMD-based GANs. 
\end{itemize}




%% file: related.tex
\section{Related Work}
\label{sec:related}

\subsection{Characteristic Kernel}
Kernel methods are widely used as a technique for developing non-linear 
algorithms. Its main idea is to map the data to a higher- or 
infinite-dimensional reproducing kernel Hilbert space (RKHS)~\cite{scholkopf2002learning}. 
\cite{gretton2008kernel} applied this idea to project the probability 
distribution to RKHS by employing linear methods to handle the high-order 
moment. A probability distribution $\mathbb{P}$ can be measured by the 
mean embedding $\mu_\mathbb{P} \in \mathcal{H}$, 
$\mu_\mathbb{P} = E_{x\sim \mathbb{P}}[\phi(x)]$, 
where $\phi(x)$ is a feature mapping function that maps the original 
data to an RKHS $\mathcal{H}$. 
Given a kernel function $k(x,x') = \langle \phi(x), \phi(x') \rangle_\mathcal{H}$, $k$ is 
characteristic iff. $\mathbb{P} \mapsto \mu_\mathbb{P}$ is injective; i.e., $\mathbb{P}$ 
is mapped to a unique element $\mu_{\mathbb{P}}$ in ${\mathcal{H}}$ \cite{fukumizu2008kernel}.

\subsection{GAN and Variants} 
Consider a sample $x$ drawn from a real dataset $\mathcal{X}$ with probability distribution 
$\mathbb{P}$. The vanilla GAN \cite{goodfellow2014generative} consists of a generator $G$ and a 
discriminator $D$. The generator $G$ maps a noise $z$ randomly sampled from a noise set 
$\mathcal{Z}$ with distribution $\mathbb{P}_z$ (e.g. a Gaussian distribution) to a probability 
distribution $\mathbb{Q}$ (i.e. $y = G(z)$ and $y \sim \mathbb{Q}$) close to $\mathbb{P}$. The 
discriminator $D$ tries to distinguish the real data $x$ from the generated data $y$, while the 
generator $G$ strives to fool the discriminator $D$ by letting $y$ as close to $x$ as possible. 
The objective function of vanilla GAN is equal to minimizing the Jensen Shannon divergence between 
$\mathbb{P}$ and $\mathbb{Q}$.



Another commonly used divergence between probability measures is integral probability 
metrics (IPM). Consider a compact space $\mathcal{X}$ in $\mathbb{R}^d$. Let $\mathscr{F}$ 
be a family of measurable and bounded real valued symmetric (i.e., $\forall f \in \mathscr{F}$, 
$-f \in \mathscr{F}$) functions on $\mathcal{X}$. Let $\mathscr{P}(\mathcal{X})$ be the set 
of measurable probability distributions on $\mathcal{X}$. Given two probability distributions 
$\mathbb{P}$, $\mathbb{Q} \in \mathscr{P}(\mathcal{X})$, the IPM indexed by the function 
space $\mathscr{F}$ is defined as follows~\cite{10.2307/1428011}:
\begin{linenomath*}
\begin{equation}
  \label{eq:IPM}
  \begin{split}
  d_\mathscr{F}(\mathbb{P},\mathbb{Q}) = \underset{f \in \mathscr{F}}{\text{sup}} \left \{  \underset{x\sim \mathbb{P}}{\mathbb{E}} f(x) - 
  \underset{x\sim \mathbb{Q}}{\mathbb{E}} f(x)\right \}.
  \end{split}
\end{equation}
\end{linenomath*}
Various distances for measuring the two distributions $\mathbb{P}$ and $\mathbb{Q}$ 
can be defined by selecting $\mathscr{F}$ appropriately \cite{sriperumbudur2012empirical}. 
Wasserstein distance considers $\mathscr{F}$ as the family of 1-Lipschitz functions (i.e. 
$\mathscr{F}:=\left \{ f \mid \left \| f \right \|_L  \leq 1\right \}$), as optimized in 
WGAN \cite{arjovsky2017wasserstein,gulrajani2017improved}. Maximum mean discrepancy (MMD) 
considers $\mathscr{F}$ as the family of functions in the unit ball in an RKHS $\mathcal{H}$ 
(i.e. $\mathscr{F}:=\left \{ f \mid \left \| f \right \|_\mathcal{H}  \leq 1\right \}$)\cite{gretton2012kernel}: 
\begin{linenomath*}
\begin{equation}
  \label{eq:MMD-}
  \begin{split}
  \text{MMD}(\mathbb{P},\mathbb{Q})
  &= \underset{\left \| f \right \|_\mathcal{H}  \leq 1}{\text{sup}} \left \{  \underset{x\sim \mathbb{P}}{\mathbb{E}} f(x) - 
  \underset{x\sim \mathbb{Q}}{\mathbb{E}} f(x)\right \} \\ 
  &= \underset{\left \| f \right \|_\mathcal{H}  \leq 1}{\text{sup}}\left \{ \left \langle f,\mu_{\mathbb{P}}-\mu_{\mathbb{Q}} \right \rangle_{\mathcal{H}} \right \} \\
  &= \underset{\left \| f \right \|_\mathcal{H}  \leq 1}{\text{sup}} \left \| \mu_{\mathbb{P}}-\mu_{\mathbb{Q}} \right \|_{\mathcal{H}}. 
  \end{split}
\end{equation}
\end{linenomath*}
\cite{gretton2012kernel} defined the squared MMD in terms of kernel $k$: 
\begin{linenomath*}
\begin{equation}
  \label{eq:squared-mmd}
  \begin{split}
  \text{MMD}^2_k(\mathbb{P},\mathbb{Q})  = & E_{x,x'\sim \mathbb{P}}\:k(x,x') - 2E_{x\sim \mathbb{P}, y\sim \mathbb{Q}}\:k(x,y) \\
   + & E_{y,y'\sim \mathbb{Q}}\:k(y,y').
  \end{split}
\end{equation}
\end{linenomath*}
\cite{gretton2012kernel} also proved that given a 
characteristic kernel $k$, $\text{MMD}^2_k(\mathbb{P},\mathbb{Q}) = 0$ iff. $\mathbb{P} = \mathbb{Q}$. 

Generative moment matching network (\gmmn) \cite{li2015generative} utilizes 
kernel and squared MMD to match all orders of statistics between a dataset and the samples generated 
by the model. \mmdgan \cite{li2017mmd} is the first GAN variant that employs MMD~\cite{gretton2012kernel}. 
\mmdgan leverages an adversarially learned kernel to improve the 
performance of \gmmn, which only uses a fixed Gaussian kernel. The discriminator of \mmdgan can be 
explained as finding an optimized kernel $\tilde{k}$ : $\tilde{k}(x,{x}')=k \circ D(x,{x}') = k(D(x),D({x}'))$ 
that can differ $\mathbb{P}$ from $\mathbb{Q}$. If $k$ is characteristic and $D$ is injective, 
then $k \circ D$ is also characteristic \cite{gretton2012kernel}. 
$\text{MMD}_{k \circ D}^2(\mathbb{P},\mathbb{Q})$ is minizied iff. $\mathbb{P}$ is equal to $\mathbb{Q}$. 
Hence the objective function of \mmdgan is defined as:
\begin{linenomath*}
\begin{equation}
  \label{eq:mmd-gan-obj}
  \underset{G}{\text{min}}\ \underset{D}{\text{max}}\ \text{MMD}^2_{k\circ D}(\mathbb{P},\mathbb{Q}).
\end{equation}
\end{linenomath*}
$k \circ D$ is approximated by an autoencoder. \mmdgan is then trained using a relaxed 
objective function based on \equationref{eq:mmd-gan-obj}.

Efforts have been made to improve the performance of \mmdgan. For example: 
in order to encourage the learning of the generator, a repulsive loss function was introduced 
in improved MMD GAN (\immdgan~\cite{wang2018improving}) to explore the details among real 
samples; SMMD GAN \cite{Arbel2018OnGR} features a principled form that imposes the Lipschitz 
constraint to the discriminator.

%% file: method.tex
\section{CKGAN}
\label{sec:ckgan}

\subsection{\ckipm}
We introduce characteristic kernel IPM (\ckipm), which considers $\mathscr{F}$ as a family of functions having the form $k(z, \Phi_{\omega}(x))$, where $k$ is a characteristic kernel. 
We will show that the IPM defined by this function is a lower bound of the MMD between the two distributions. Given a characteristic kernel $k(x, x') = \langle \phi(x), \phi(x') \rangle_\mathcal{H}$, we consider the following function family: 
\begin{align*}
     & \mathscr{F} := \{ f(x) = \langle \phi(z), \phi (\Phi_{\omega}(x))  \rangle  \mid z \in \mathbb{R}^m, \\ 
     & \Phi_{\omega}: \mathcal{X} \rightarrow \mathbb{R}^m, \omega \in \Omega  \}, 
\end{align*}
where $\phi(\cdot)$ is a feature mapping function that maps the original data to an RKHS $\mathcal{H}$, and $\Phi_{\omega}$ is a parameterized neural network as a non-linear feature mapping from $\mathcal{X}$ to $\mathbb{R}^m$. 

Moreover, we consider characteristic kernels satisfying $\left \| \phi(\cdot) \right \|_\mathcal{H} \leq 1$. This constraint corresponds to the family of functions in the unit ball in $\mathcal{H}$ for MMD~\cite{gretton2012kernel}~\footnote{The MMD framework requires the kernel to be uniformly bounded by some constant, not strictly by 1. Here, we consider the unit ball case for simplicity, in line with \cite{gretton2012kernel}. For the case when the kernel of MMD is bounded by a constant, the $\phi$ function of the characteristic kernel can be bounded accordingly. This applies to the RBF mixture and Matern kernels used in our experiments.}. 
Then for the \ckipm $d_\mathscr{F}(\mathbb{P},\mathbb{Q})$, we have
\begin{equation}
  \left \| f(x) \right \|^2_\mathcal{H} = \left \| k(z, \Phi_\omega(x)) \right \|^2_\mathcal{H} = k(z, z) = \langle \phi(z), \phi(z) \rangle \leq 1. 
\end{equation}
Therefore, $\left \| f \right \|_\mathcal{H} \leq 1$, and $\mathscr{F} \subseteq \{ f \in \mathcal{H} : \left \| f \right \|_\mathcal{H} \leq 1 \}$. Because the supremum taken over a subset of functions is always less than or equal to the supremum taken over a larger set, we have 
\begin{equation}
  d_\mathscr{F}(\mathbb{P}, \mathbb{Q}) = \sup_{f\in\mathscr{F}} \Big\{ \underset{x\sim \mathbb{P}}{\mathbb{E}}[f(x)] - \underset{x\sim \mathbb{Q}}{\mathbb{E}}[f(x)] \Big\} \le \sup_{f\in \{f\in\mathcal{H}:\|f\|_\mathcal{H}\le1\}} \Big\{ \underset{x\sim \mathbb{P}}{\mathbb{E}}[f(x)] - \underset{x\sim \mathbb{Q}}{\mathbb{E}}[f(x)] \Big\} = \text{MMD}(\mathbb{P},\mathbb{Q}). 
\end{equation}
We can see that \ckipm is a lower bound of MMD. 
A GAN based on \ckipm aims to optimize this lower bound. 

\subsection{Training GAN with \ckipm}
By incorporating the above \ckipm, we design the objective function of 
\ckgan as the following adversarial game:
\begin{linenomath*}
\begin{equation}
  \label{eq:ckgan-obj}
  \underset{G}{\text{min}}\ \underset{D}{\text{max}}\ d_{\mathscr{F}}(\mathbb{P},\mathbb{Q}), 
\end{equation}
\end{linenomath*}
where
\begin{linenomath*}
\begin{align*}
d_{\mathscr{F}}(\mathbb{P},\mathbb{Q}) = \underset{x \sim \mathbb{P}, z \sim \mathbb{P}_z}{\mathbb{E}}k(z, D(x)) - \underset{z \sim \mathbb{P}_z}{\mathbb{E}}k(z, D(G(z))). 
\end{align*}
\end{linenomath*}

To use the kernel $k$ along with the discriminator $D$, i.e., $k \circ D(x,{x}') = k(D(x),D({x}'))$, 
we expect $D$ to be injective~\cite{li2017mmd}, i.e., there exists an function $D^{-1}$ such 
that $D^{-1}(D(x)) = x, {\forall} x\in \mathcal{X}$ and $D^{-1}(D(G(z'))) = G(z')$, 
${\forall} z\in \mathcal{Z}$. In \mmdgan, the injection is approximated by using an autoencoder. In 
\ckgan, we approximate the injection by mapping the generated images back to the random noise. 

The kernel function plays a role of a similarity measure between the noise and the mapping output. 
Intuitively, the discriminator $D$ wants to reduce the similarity between the noise $z$ and the 
embedding of the generated data $G(z)$ and increase the similarity between $z$ and the embedding of 
the real data $x$, while the generator $G$ wants to increase the similarity between $z$ and the 
embedding of $G(z)$. This training strategy helps \ckgan to alleviate the mode collapse 
problem. This is because when mode collapse happens, $G(z)$ falls to a narrow region compared to the 
real data distribution, and then $D(G(z))$ will significantly differ from $z$. This will be penalized 
by the $k(z, D(G(z)))$ term and result in a large loss. To train \ckgan, we also apply the gradient penalty~\cite{Arbel2018OnGR} to 
ensure the function set $\mathscr{F}$ is bounded and stabilize the training process. 

We also note that in \equationref{eq:ckgan-obj}, the $k(z, D(G(z)))$ term is similar to the reconstruction loss in \veegan but there 
are two differences: First, we regard the discriminator as an encoder instead of introducing an extra 
reconstructor network. Second, we use a kernel function rather than an $L_2$ 
loss as the similarity measure between $z$ and $D(G(z))$. The reason is that the 
$L_2$ loss will use up almost all the degrees of the freedom of $D(G(z))$~\cite{su2019gan}, 
leaving little room for the discriminator to distinguish the real and the 
generated data. In contrast, a kernel function yields more freedom and enables 
the GAN to generate images of higher quality.

\subsection{Kernel Function Selection}
\label{sec:select}
There are many options for the characteristic kernel $k$, e.g., Gaussian kernel, 
Laplacian kernel, and Exponential kernel. \mmdgan resorts to an adversarially 
learned kernel but only considers the case of a mixture (i.e., a sum) of Gaussian 
(RBF) kernels with different standard deviations. For \ckgan, we explore in the 
direction of incorporating a variety of characteristic kernels. Since it is 
time-consuming to manually tune a kernel function for a specific task, we propose 
to learn a characteristic kernel $k^{lc}$ to save this effort. 

Our solution is based on a soft selection of $K$ characteristic kernels. We consider 
$k^{lc}$ as a linear combination of characteristic kernels, whose weights are learned 
together with the GAN. We define $k^{lc}$ as follows.
\begin{linenomath*}
\begin{equation}
    \label{eq:select-linear}
    k^{lc}(x, x') = \sum_{i=1}^{K} \xi_i k_i(x, x'),   
\end{equation}
\end{linenomath*}
A property is that a linear combination of characteristic kernels is also characteristic 
\cite{sriperumbudur2010hilbert}. So $k^{lc}$ is characteristic and can be used as the 
kernel $k$ in \equationref{eq:ckgan-obj}. 

To concentrate on the kernels that have more effect on achieving the optimum for 
training, we apply a $\softmax$ function to obtain the weights. Consider a weight 
vector $\mathbf{\xi}' = [\xi'_1, \ldots, \xi'_K]$, we have 
$\mathbf{\xi} := \softmax(\mathbf{\xi}')$ where $\mathbf{\xi}'$ is determined 
through training. 

Other kernel selection options include 
\begin{inparaenum} [(1)]
  \item using a linear combination directly without the $\softmax$, i.e., 
  $k^{lc}(x, x') = \sum_{i=1}^{K} \xi_i k_i(x, x')$, $\xi_i \in [0, 1]$, and $\sum_{i=1}^{K}\xi_i = 1$; 
  \item using a one-hot vector that follows the $\softmax$, i.e., $k^{lc}(x, x') = k_j(x, x')$, 
  $j = \argmax_i \xi_i$. 
\end{inparaenum}
The advantage of soft selection over these options are: 
\begin{inparaenum} [(1)]
  \item Soft selection puts more weights on important kernels, and thus the 
  training process converges faster than using a linear combination directly. 
  \item The one-hot method chooses only one function as $k$, while soft selection 
  leverages the capability of all the $K$ component kernels. 
\end{inparaenum}


Algorithm~\ref{alg:ckgan} provides the pseudo-code of training \ckgan based on 
RMSProp~\cite{hinton2012lecture}. The default parameter setting is also given. 
We assume $k = k^{lc}$ in the pseudo-code. The weight vector $\mathbf{\xi}$ 
is updated twice in each iteration because the loss of the generator is also useful 
in updating the kernel weights. When a manually tuned kernel is used for $k$, we 
can omit the $\mathbf{\xi}$ vector. The time complexity of each iteration is 
$O(KB)$, where $K$ is the number of component kernels in the $\mathbf{\xi}$ vector and 
$B$ is the batch size. In contrast, the time complexity of \mmdgan is 
$O(KB^2)$~\cite{li2017mmd}, where $K$ is the number of kernels in the RBF mixture. 

In Algorithm~\ref{alg:ckgan}, the learning of kernel weights alternates between the gradient 
steps on the discriminator loss $L_D$ and the gradient steps on the generator loss $L_G$. We 
have considered three settings for kernel weight learning: (1) learn the kernel weight by 
only optimizing $L_G$, (2) by only optimizing $L_D$, and (3) by optimizing both. Our 
experimental results showed that method (3) is slightly better than the other two. Thus, we 
choose to learn the kernel weights by optimizing both $L_D$ and $L_G$. We believe that the 
advantage of this update strategy is that the kernel weights play a role in the adversarial 
game between $D$ and $G$. In each iteration of training, $D$ finds the best kernel weight 
to separate real and generated data. After that, $G$ tries to fool $D$ by not only changing 
its own parameters but also the kernel weights. In doing so, $D$'s ability of 
discriminating real \& fake data and $G$'s ability of generating fake data are both 
improved. 

\begin{algorithm}[t]
    \Input{gradient penalty coefficient $\lambda$, characteristic kernel functions $k_1, \ldots, k_K$, learning rate $\eta$, the number of discriminator iterations per generator iteration $n_{d}$, and batch size $B$.}
    \State{initialize generator parameters $\theta$, discriminator parameters $\omega$, and kernel 
    selection parameters $\xi$} \label{ln:initialize}
    \While{$\theta$ has not converged}{
        \For{$t = 1, \ldots, n_{d}$}{
            \For{$i = 1, \ldots, B$}{ \label{ln:train-discriminator-begin}
                \State{sample real data $x \sim \mathbb{P(\mathcal{X})}$, latent variables $z \sim \mathbb{P(\mathcal{Z})}$}
                \State{$L_D^{(i)} \gets k(z, D(G(z))) -  k(z,D(x))$ $+ \lambda\left \|\nabla_{\hat{x}} D(\hat{x})  \right \|_2^2$} 
            }
            \State{$\omega$ $\gets$ $\omega$ - RMSProp($\nabla_{\omega}\frac{1}{B}\sum_{i}^{B}L_D^{(i)}$, $\omega$, $\eta$)} \label{ln:train-discriminator-end}
            \State{$\xi$ $\gets$ $\xi$ - RMSProp($\nabla_{\xi}\frac{1}{B}\sum_{i}^{B}L_D^{(i)}$, $\xi$, $\eta$)} \label{ln:update-weight-first}
        }
        \State{sample a minibatch $\{z\}_{i=1}^B \sim \mathbb{P(\mathcal{Z})}$} \label{ln:train-generator-begin}
        \State{$L_G \gets -k(z, D(G(z)))$}
        \State{$\theta$ $\gets$ $\theta$ - RMSProp($\nabla_{\theta}\frac{1}{B}\sum_{i}^{B}L_G^{(i)}$, $ \theta$, $\eta$)} \label{ln:train-generator-end}
        \State{$\xi$ $\gets$ $\xi$ - RMSProp($\nabla_{\xi}\frac{1}{B}\sum_{i}^{B}L_G^{(i)}$, $ \xi$, $\eta$)} \label{ln:update-weight-second}
    }
    \caption{CKGAN (default $\lambda=10$, $\eta=0.00005$, $n_{d}=5$, $B=64$)}    
    \label{alg:ckgan}    
\end{algorithm}

%% file: relation.tex
\section{Relationship to Other GAN Variants}
\label{sec:relationship}
\begin{inparaenum} [(1)]
  \item \ckgan optimizes the lower bound of MMD, while the \mmdgan family 
  optimizes the squared MMD. Unlike \mmdgan, the discriminator $D$ in 
  \ckgan is composed of an encoder rather than the combination of an encoder 
  and a decoder. Such design of \ckgan can save computing resources. 
  \item If we consider $\mathscr{F}$ as the set of 1-Lipschitz functions, use 
  linear kernel rather than characteristic kernel, and constrain the output 
  dimension of discriminator to be 1, then our loss function will resemble that 
  of WGAN~\cite{arjovsky2017wasserstein}. WGAN can be described as matching the 
  first-order statistics, while \ckgan tries to match the high-order moments 
  using kernel tricks. 
  \item \veegan establishes an extra reconstructor network $E$ to map the 
  generated sample $G(z)$ to a latent code $E(G(z))$ and then minimizes the 
  $L_2$ loss between $z$ and $E(G(z))$. \ckgan regards the discriminator $D$ 
  as an encoder and the generator $G$ as a decoder to approximate the 
  reconstruction error by minimizing the similarity between $z$ and $D(G(z))$. 
\end{inparaenum}

%% file: exp.tex
\section{Experiments}
\label{sec:exp}

We report the most important experimental results here. Please see 
\ref{sec:exp-setup} for experiment setup details and \ref{sec:additional-exp} 
for additional experiments, including more generated samples with 
higher resolution, the effect of kernel parameters, the average 
generation time and \ckgan with spectral normalization. 

\begin{table*}[t]
  \small
  \centering
  \setlength{\tabcolsep}{1.94mm}{
  \caption{The number of modes (\#modes) captured (higher is better), the percentage of high-quality 
  (\%HQ) samples (higher is better), KL divergence (lower is better), and training time (in seconds) 
  on synthetic datasets. Results are averaged over 10 trials.} 

  \label{tab:table_Synthetic}
  \centering
  \resizebox{\linewidth}{!}{
  \begin{tabular}{lcccccccccccccccc} 
    \toprule
    & \multicolumn{4}{c}{2D Ring} & \multicolumn{4}{c}{2D Grid} & \multicolumn{4}{c}{2D SmileFace} \\
    \cmidrule(r){2-5} \cmidrule(r){6-9} \cmidrule(r){10-13}
    & \#modes & \multirow{2}{*}{\%HQ} & \multirow{2}{*}{KL} & \multirow{2}{*}{Time} & \#modes & \multirow{2}{*}{\%HQ} & \multirow{2}{*}{KL} & \multirow{2}{*}{Time} & \#modes & \multirow{2}{*}{\%HQ} & \multirow{2}{*}{KL} & \multirow{2}{*}{Time}\\
    & (max=8) & & & & (max=25) & & & & (max=2) &  &&\\
    \midrule
    \unrolledgan &8	&97.13&	0.0167&	61.9&	22&	87.48&	0.0936&	62.95&	0&	0&	1.0849&	73.35\\
    \veegan &8	&97.82	&0.0230	&27.47	&25	&98.75	&0.0765	&27.84	&2	&98.42	&0.0026	&32.90     \\
    \mmdgan &8	&90.09	&0.0025	&47.88	&25	&96.42	&\textbf{0.0052}	&49.74	&2	&74.71	&0.0010	&261.15     \\
    \smmdgan &8	&94.21	&0.0032	&32.97	&25	&94.78	&0.0064	&34.61	&2	&59.3	&0.0138	&170.47    \\
    \immdgan &8	&98.03	&\textbf{0.0021}	&32.54	&25	&94.99	&0.0057	&34.04	&2	&97.85	&\textbf{0.0007}	&167.30     \\

    \midrule
    \ckgan $k^g$  &8	&98.07	&\textbf{0.0021}    &17.89	&25	&98.64	&0.0054	&18.14	&2	&98.33	&0.0103	&21.58    \\
    \ckgan $k^l$  &8	&98.33	&0.0060	&17.89	&25	&97.17	&0.0119	&\textbf{18.09}	&2	&  98.26	&0.0013	&\textbf{21.54}    \\
    \ckgan $k^{rbfm}$  &8	&98.50	&0.0025	&\textbf{17.86}	&25	&\textbf{99.08}	&0.0058	&18.17	&2	&98.70	&0.0010	&21.62    \\
    \ckgan $k^{e}$   &8	&98.52	&0.0051	&17.87	&25	&98.44	&0.0080	&18.19	&2	&98.95	&0.0033	&21.57   \\
    \ckgan $k^{m 3/2}$  &8	&98.34	&0.0025	&17.87	&25	&97.24	&0.0077	&18.15	&2	&98.01	&0.0009	&21.62    \\
    \ckgan $k^{m 5/2}$  &8	&\textbf{98.95}	&0.0029	&17.95	&25	&96.46	&0.0076	&18.22	&2	&\textbf{98.96}	&0.0015	&21.66    \\
    \ckgan $k^{lc}$ &8	&98.46	&0.0037	&19.06	&25	&98.98	&0.0077	&19.27	&2	&98.57	&0.0011	&22.73    \\
    
    \bottomrule
  \end{tabular}
  }
  }
\end{table*}

\begin{figure*}[t]
  \centering
  \includegraphics[width=1\textwidth]{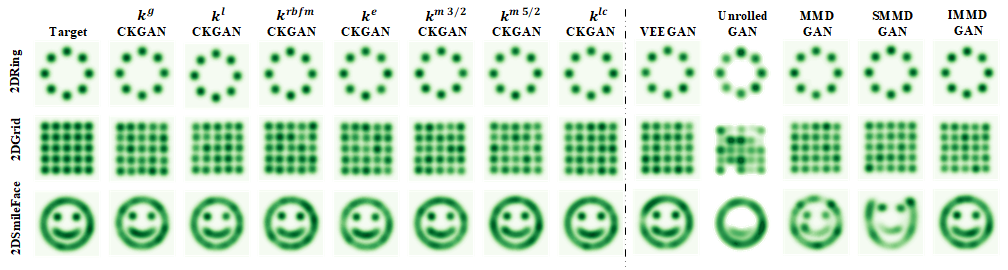}  
  \caption{Density plots of target distribution (real data) and distributions generated by GAN variants.} 
  \label{fig:synthetic_datasets_generation}
\end{figure*}

\begin{figure*}[t] 
  \centering
  \subfigure[\ckgan $k^{g}$]{
    \includegraphics[width=5.7cm,height=5.8cm]{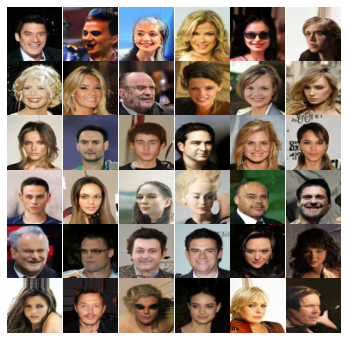}
    \label{fig:qualitative_analysis_CKGAN_gaussian_generation}
  }
  \hspace{-1.0mm}
  \subfigure[\ckgan $k^{lc}$]{
  \includegraphics[width=5.7cm,height=5.8cm]{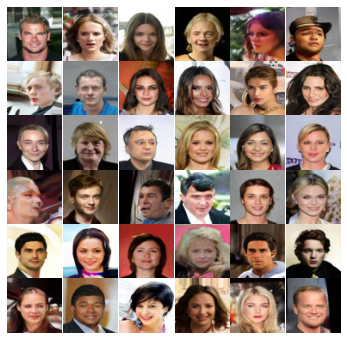}
  \label{fig:qualitative_analysis_CKGAN_lk_generation}
  }
  \subfigure[\veegan]{
  \includegraphics[width=5.7cm,height=5.8cm]{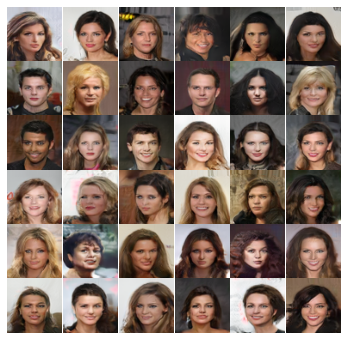}
  \label{fig:qualitative_analysis_VEEGAN_generation}
  }
  \hspace{-1.0mm}
  \subfigure[\mmdgan]{
    \includegraphics[width=5.7cm,height=5.8cm]{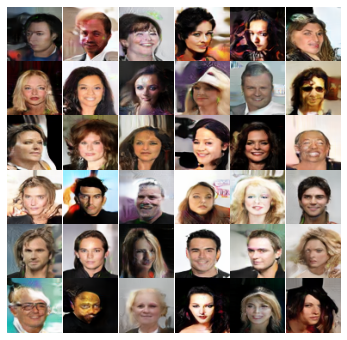}
  }
  \hspace{-1.0mm}
  \subfigure[\smmdgan]{
    \includegraphics[width=5.7cm,height=5.8cm]{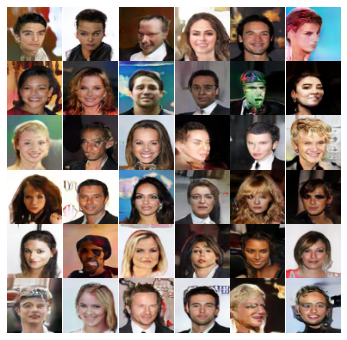}
  }
  \subfigure[\immdgan]{
    \includegraphics[width=5.7cm,height=5.8cm]{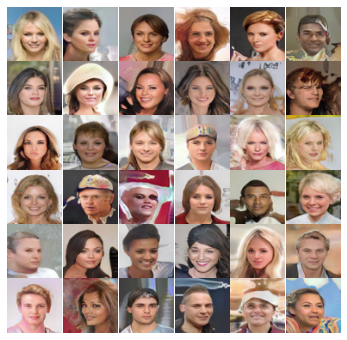}
    \label{fig:qualitative_analysis_IMMD_generation}
  }
  \hspace{-1.0mm}
  \caption{Randomly generated samples by GAN variants on CelebA.}
  \label{fig:qualitative_analysis_generation}
\end{figure*}

We conduct experiments on synthetic and real image datasets. The synthetic datasets are 
\textbf{2D Ring} \cite{srivastava2017veegan}, \textbf{2D Grid} \cite{srivastava2017veegan}, 
and \textbf{2D SmileFace} synthesized by ourselves. The real image datasets are 
\textbf{MNIST} of $28 \times 28$ handwritten digits \cite{lecun1998gradient}, 
\textbf{CIFAR-10} of $32 \times 32$ images \cite{krizhevsky2009learning}, \textbf{CelebA} 
images cropped and resized to $64 \times 64$ \cite{liu2015deep}, and \textbf{LSUN} church 
outdoor pictures resized to $64 \times 64$ \cite{yu2015lsun}. We equip \ckgan with seven 
characteristic kernel options: Gaussian $k^g$, Laplacian $k^l$, RBF mixture $k^{rbfm}$, 
exponential $k^{e}$, Matern 3/2 $k^{m3/2}$, Matern 5/2 $k^{m5/2}$, and a soft selection 
$k^{lc}$ over the other six kernels. The first six options are manually tuned for best 
overall performance. $k^{lc}$ is described in Section~\ref{sec:select}. 
We compare with \unrolledgan~\cite{metz2016unrolled}, \veegan~\cite{srivastava2017veegan}, 
and the \mmdgan family (including \mmdgan~\cite{li2017mmd}, \smmdgan~\cite{Arbel2018OnGR}, 
and \immdgan~\cite{wang2018improving}). 

\subsection{Evaluation on Synthetic Datasets}
We first measure \ckgan's capability of mitigating mode collapse and compare with 
other GAN variants. For fair comparison, we use the same architecture (see Appendix~\ref{sec:network-arc}) 
in generator and discriminator for all the GANs. We test on synthetic datasets, which are 
suitable for evaluating the degree of mode collapse since their distributions and the 
numbers of modes are predefined. We measure four metrics (see Appendix~\ref{sec:metrics} 
for definitions): 
the number of modes captured, the percentage of high-quality samples, the KL divergence, 
and the training time. 
The quantitative results of the competitors are reported in Table~\ref{tab:table_Synthetic}. 
The visualization results are shown in Figure~\ref{fig:synthetic_datasets_generation}. 

The quantitative comparison shows that \ckgan captures all the modes. When equipped 
with the automatically selected kernel, it consistently outperforms \veegan and 
the \mmdgan family by producing more percentage of high-quality samples, and is 
competitive with them in terms of KL divergence. Note that, since \unrolledgan fails 
to capture the ``eye'' mode on 2D SmileFace, its mode number and KL metrics are zero. 
\ckgan is also the fastest in training. The competitors in the \mmdgan family are slow 
because they need to compute the kernel matrix and Jacobian matrix in each iteration; 
both procedures are time-consuming, especially when the number of training data 
increases. For \ckgan, we also observe that none of the kernel options dominates others 
in all the cases. Although the performance of the automatically selected kernel is in 
the middle of the seven options, we will see better results on real images tasks. Around 
6\% additional training time is required to obtain the automatically selected kernel. 
This is acceptable because to find the best of the other six kernels, we need to manually 
tune and try all of them.


The visualization results show that \ckgan performs significantly better than 
\mmdgan and \smmdgan on 2D SmileFace. Although \immdgan produces similar excellent 
results to \ckgan's, we will see the difference in the real image tasks. 
To highlight the visual difference from \veegan, we also devise an simpler 
architecture where \ckgan performs clearly better 
(Appendix~\ref{sec:More_Synthetic_datasets_Experiments}). 
In sum, the results on synthetic datasets demonstrate the strong reliability and 
generality of \ckgan.

\begin{table*}[!t]
  \small
  \centering
  \setlength{\tabcolsep}{2.2mm}{
  \caption{IS (higher is better) and FID (lower is better)  on real image datasets. Results are averaged over 5 trials.} 
  \label{tab:table_small_images_generation_metrics}
  \centering
  \resizebox{\linewidth}{!}{
    \begin{threeparttable}
      \begin{tabular}{lcccccccccc}
        \toprule
    
      \multirow{2}{*}{Methods} &\multicolumn{2}{c}{MNIST}&\multicolumn{2}{c}{CIFAR-10}&\multicolumn{2}{c}{CelebA}&\multicolumn{2}{c}{LSUN}             \\
        \cmidrule(r){2-3} \cmidrule(r){4-5} \cmidrule(r){6-7} \cmidrule(r){8-9} 
         & IS & FID & IS & FID & IS & FID & IS & FID  \\%
      \midrule
      Real data & 2.56$\pm$0.03 & 0.48$\pm$0.01 & 11.29$\pm$0.15 & 1.65$\pm$0.01 & 3.66$\pm$0.04 & 2.64$\pm$0.01 & 3.12$\pm$0.04 & 2.63$\pm$0.01\\
      \midrule
        
      \veegan       &2.55$\pm$0.03	&3.96$\pm$0.25 	&6.68$\pm$0.21	&30.57$\pm$1.78	&2.77$\pm$0.05	&15.07$\pm$0.72	 &2.67$\pm$0.04	&12.95$\pm$0.81	\\
      \mmdgan   &2.44$\pm$0.02	&8.81$\pm$0.19 	&6.05$\pm$0.04	&41.27$\pm$0.75	&2.90$\pm$0.02	&15.83$\pm$0.18	  &2.90$\pm$0.04	&17.18$\pm$0.57	\\
      \mmdgan $k^{il}$    	&2.45$\pm$0.02	&8.67$\pm$0.52	&6.70$\pm$0.08	&38.76$\pm$0.83	&2.80$\pm$0.02	&40.75$\pm$1.21		&2.83$\pm$0.10	&55.00$\pm$0.52  \\
      \mmdgan $k^{lc}$    &2.46$\pm$0.02	&8.15$\pm$0.36	&6.20$\pm$0.09	&36.83$\pm$0.72      &2.95$\pm$0.04	 &13.07$\pm$0.19	       &2.66$\pm$0.05 	&16.61$\pm$0.79 	\\
      \midrule
      \smmdgan  &2.49$\pm$0.06	&3.44$\pm$0.06 	&7.24$\pm$0.13	&25.49$\pm$0.43     &2.95$\pm$0.03	&17.31$\pm$0.53	  &2.83$\pm$0.03	&12.75$\pm$0.46    \\
      \smmdgan $k^{il}$	&2.52$\pm$0.03	&3.46$\pm$0.25	&7.35$\pm$0.09	&25.36$\pm$0.66	&2.99$\pm$0.04	&16.37$\pm$0.70	&2.90$\pm$0.03	&12.68$\pm$0.14  \\
      \smmdgan $k^{lc}$  &2.51$\pm$0.03	&3.37$\pm$0.01	&7.26$\pm$0.10	&24.13$\pm$0.14      &3.02$\pm$0.07	&16.16$\pm$0.50	  &2.86$\pm$0.04	&12.32$\pm$0.61	\\
      \midrule
      \immdgan   &2.52$\pm$0.02	&3.32$\pm$0.18	&7.55$\pm$0.02	&22.33$\pm$0.61     &3.09$\pm$0.07	&12.78$\pm$1.24	  &\textbf{2.93$\pm$0.06}	&8.34$\pm$0.61	\\
    \immdgan $k^{il}$	&2.45$\pm$0.02	&3.04$\pm$0.08	&\textbf{7.77$\pm$0.12}	&21.50$\pm$0.87	&3.04$\pm$0.10     &11.99$\pm$0.15	&2.90$\pm$0.09	&8.32$\pm$0.48	 \\
      \immdgan $k^{lc}$   &2.49$\pm$0.06	&3.19$\pm$0.06	&7.76$\pm$0.04	 &21.34$\pm$0.71    &\textbf{3.13$\pm$0.05}	&11.85$\pm$0.25	&2.88$\pm$0.05	&8.14$\pm$0.37\\
    
      \midrule
    
      \ckgan $k^g$            &2.55$\pm$0.02       &2.76$\pm$0.15         &7.72$\pm$0.07	&20.48$\pm$0.40	&3.07$\pm$0.02	&7.17$\pm$0.31	   &2.78$\pm$0.04	&8.11$\pm$0.33	\\
      \ckgan $k^l$             &2.50$\pm$0.02	&2.79$\pm$0.21        &7.65$\pm$0.08	&20.87$\pm$0.59 	&3.04$\pm$0.05	&7.02$\pm$0.33	  &2.76$\pm$0.02	&8.05$\pm$0.25	\\
      \ckgan $k^{rbfm}$    &\textbf{2.56$\pm$0.02}	&2.86$\pm$0.10        &7.74$\pm$0.09	&20.30$\pm$0.10 	&3.02$\pm$0.03	&7.38$\pm$0.20		&2.76$\pm$0.03	&8.83$\pm$0.59	\\
      \ckgan $k^{e}$         &2.51$\pm$0.02	&2.78$\pm$0.10        &7.54$\pm$0.06	&21.10$\pm$0.23 	&3.09$\pm$0.03	&\textbf{6.62$\pm$0.25}		&2.82$\pm$0.04	&\textbf{7.73$\pm$0.48}	\\
      \ckgan $k^{m3/2}$   &2.52$\pm$0.02	&2.90$\pm$0.15        &7.68$\pm$0.05	&\textbf{20.02$\pm$0.23} 	&3.11$\pm$0.07	&6.74$\pm$0.20		&2.85$\pm$0.05	 &8.02$\pm$0.46	\\
      \ckgan $k^{m5/2}$   &2.53$\pm$0.03	&2.75$\pm$0.25       &7.66$\pm$0.04	&21.03$\pm$0.58	&3.09$\pm$0.03	&6.68$\pm$0.37	 	&2.80$\pm$0.04	 &8.08$\pm$0.57	\\

      \ckgan $k^{lc}$        &2.53$\pm$0.02	&\textbf{2.57$\pm$0.07}       &7.75$\pm$0.05	&20.80$\pm$0.98	 &3.10$\pm$0.05	&6.95$\pm$0.54		&2.76$\pm$0.09	&7.87$\pm$0.34	\\
      \bottomrule

      \end{tabular}
  \end{threeparttable}
  }
  }
\end{table*}

\subsection{Evaluation on Real Image Datasets}
\label{sec:exp-real-datasets}
For fair comparison, we use the same architecture (see Appendix~\ref{sec:network-arc}) 
in generator and discriminator for all the GANs. Note that this may result in the 
reported results differ from what was reported in previous studies (nonetheless, we 
observe some results even better than previously reported; e.g., \smmdgan's inception 
score on CelebA). 

\myparagraph{Qualitative Analysis}
\label{section:qualitative_analysis}
Figure~\ref{fig:qualitative_analysis_generation} shows the generated samples 
produced by the competitors trained on the CelebA dataset. \ckgan generates 
high-quality and diverse samples. By comparing the hairstyle, skin, light exposure, 
and the background, we find \veegan and \immdgan are more likely to generate 
female faces. \immdgan also tends to generated images with light color. 
\mmdgan and \smmdgan generated twisted faces.

\myparagraph{Quantitative Analysis}
Table~\ref{tab:table_small_images_generation_metrics} reports the quantitative results 
on the real image datasets.
To evaluate the visual quality and diversity of generated images, 
we measure two metrics (see Appendix~\ref{sec:metrics} for definitions): Inception Score 
(IS) and Fr\'{e}chet inception distance (FID). To 
investigate the effectiveness of the proposed kernel function selection techniques 
($k^{lc}$), we also apply it to the competitors in the \mmdgan family. 

In terms of IS, \ckgan equipped with $k^{lc}$ performs better than \veegan on all 
the datasets except MNIST. It also outperforms the original version of the \mmdgan family 
except on LSUN. This shows the capability of \ckgan in mitigating the mode collapse 
problem. In terms of FID, \ckgan equipped with $k^{lc}$ always performs better than other 
GANs. 
This suggests that optimizing the lower bound of MMD is effective in learning a data 
distribution. When the proposed kernel is employed to the \mmdgan family, IS is 
generally improved except in a few cases, and FID is improved in all the cases. This 
means the proposed kernel selective strategy is also helpful to other MMD-based GANs and 
is not limited to a specific objective function. 

\begin{table}[!t]
  \small
  \centering
  \caption{Training time (in seconds, 1000 iterations) on real image datasets. } 
  \label{tab:table_real_dataset_training_time}
  \centering
      \begin{tabular}{lcccccccccc}
        \toprule
        Methods & MNIST & CIFAR-10 & CelebA & LSUN \\
        \midrule

        \mmdgan	 &93.87	 &129.31	 &309.53	 &309.71\\
        \smmdgan	 &43.96	 &\textbf{52.57}	 &378.95	 &373.92\\
        \immdgan	 &145.66	 &175.18	 &484.02	 &479.47\\
        \midrule				
        \ckgan $k^g$	 &43.38	 &57.62	 &157.27	 &160.32\\
        \ckgan $k^l$ 	 &\textbf{42.06}	 &60.26	 &\textbf{156.90}	 &159.32\\
        \ckgan $k^{rbfm}$	 &45.63	 &64.66	 &161.84	 &\textbf{157.64}\\
        \ckgan $k^{e}$	 &43.41	 &58.36	 &162.92	 &161.59\\
        \ckgan $k^{m3/2}$	 &44.29	 &63.85	 &162.67	 &162.01\\
        \ckgan $k^{m5/2}$	 &45.03	 &59.79	 &162.00	 &158.88\\
        \ckgan $k^{lc}$	 &55.64	 &67.52	 &164.43	 &163.15\\
        \bottomrule

      \end{tabular}
\end{table}

In addition, we observe that the performance of the automatically selected kernel $k^{lc}$ 
roughly approaches the best of manually tuned ones, and achieves the best in IS on CIFAR-10 
and FID on MNIST. None of the kernel option dominates others in all the cases. We also report 
training time in Table~\ref{tab:table_real_dataset_training_time}. Compared to manually 
selected kernels, the automatically selected one spends slightly more training time: on average 
26\% on MNIST, 11\% on CIFAR-10, 2\% on CelebA and LSUN. The results indicate that \ckgan is 
able to incorporate various characteristic kernels and saves us from manually choosing one out 
of many available options. The automatically selected kernel is tailored to the dataset and 
has strong fault tolerance; i.e., it combines the advantages of multiple kernels and is less 
vulnerable to the case when some kernel does not deliver good performance on a specific dataset 
(e.g., RBF mixture kernel reports relatively high FID on MNIST, CelebA, and LSUN).

\myparagraph{Comparison with Implicit Learned Kernel}
We compare the characteristic kernel selected by our method ($k^{lc}$) with implicit learned 
kernel ($k^{il}$) ~\cite{li2019implicit} by applying the two methods to \mmdgan, \smmdgan, and 
\immdgan, respectively. Table~\ref{tab:table_small_images_generation_metrics} reports IS and FID on real image 
datasets. We observe that in most test cases, both kernel methods perform better than the 
original version of the three GAN variants. For IS, implicit learned kernel reports higher score than 
our kernel on CIFAR-10 and LSUN, but ours is better on the other two datasets except for 
\smmdgan on MNIST. For FID, our kernel prevails in 11 out of 12 test cases. We also show the 
results of \ckgan for comparison. Note that implicit learned kernel is not available for \ckgan 
because implicit learned kernel computes a kernel matrix, which is missing in \ckgan. We observe 
that our kernel is more effective when equipped on \ckgan, which outperforms all the others in 
terms of FID and delivers the best IS on MNIST.

%% file: concl.tex
\section{Conclusion}
\label{sec:concl}

We proposed \ckgan, a general GAN variant with a \ckipm framework to learn a 
distribution. We show that \ckipm is a lower bound of MMD in a reproducing 
kernel Hilbert space and \ckgan aims to optimize this lower bound. By mapping 
the generated images back to random noise, \ckgan is able to alleviate the 
mode collapse problem. We also considered a variety of characteristic kernel 
options and proposed a soft selection method to learn a characteristic kernel 
together with the GAN. Our experiments demonstrated that \ckgan generates 
images with high quality and diversity, and outperforms other GAN variants. 
The performance of \ckgan with the soft selection is very close to the best 
of manually fine-tuned ones and saves the effort in tuning the kernel 
function. In addition, the soft selection can be also used to enhance other 
MMD-based GAN variants. 


%% file: ack.tex
\section*{Acknowledgement}
This work was supported by JSPS Kakenhi JP17H06099, JP18H04093, and JP19K11979, JP23K17456, JP23K25157, JP23K28096, and JST CREST JPMJCR22M2.

%% file: appendix.tex
\section*{Appendix}

\section{Experiment Setup}
\label{sec:exp-setup}

\subsection{Datasets}
\label{sec:exp-setup-datasets}
To evaluate the effectiveness of \ckgan, we experiment on synthetic datasets 
and real image datasets. The synthetic datasets are a mixture of eight 2D 
Gaussian distributions arranged in a ring (2D Ring), a mixture of twenty-five 
2D Gaussian distributions arranged in a grid (2D Grid), and a mixture of a 
circle surrounding half of an elliptical placed in the mouth part and two 
Gaussian placed in the eye part of the human face (2D SmileFace). For 2D Ring 
and 2D Grid, we follow the synthetic strategies of PacGAN~\cite{lin2018pacgan} 
except for setting the standard deviation to 0.0001 for 2D Ring and to 0.005 
for 2D Grid. For 2D SmileFace, the input dataset is generated by a 
mixture of: (1) two Gaussians with means (-0.4, 0.3) and (0.4, 0.3) and 
variances 0.001, and (2) half of an ellipse whose semi-major axis is 0.6 and 
semi-minor axis is 0.5 on a circle with a radius of 1. The real image datasets 
have been introduced in Section~\ref{sec:exp}. 

\subsection{Kernel Functions}
\label{sec:kernel-func}
We equip \ckgan with the following characteristic kernels: 

\begin{enumerate}
  \item Gaussian kernel: $k^g(x,y) = \exp \left ( -\frac{\left \| x-y \right \|_2^2}{2\sigma^2} \right )$.
  \item Laplacian kernel: $k^l(x,y) = \exp \left ( -\frac{\left \| x-y \right \|_1}{\sigma} \right ) $.
  \item RBF mixture kernel: $k^{rbfm}(x,y) = \sum_{q}^{K} k_{\sigma_q}^g(x,y)$ 
  where $k^g_{\sigma_q}$ is a Gaussian kernel with the parameter $\sigma_q$. In our work, we follow the 
  parameters suggested for \mmdgan~\cite{li2017mmd} to set $K = 5$ and use $\set{1, 2, 4, 8, 16}$ as the 
  five $\sigma_q$ values.
  \item Exponential kernel: $k^{e}(x,y)= \exp(-\frac{\left \| x-y \right \|_2}{\sigma})$ \cite{sriperumbudur2010hilbert}.
  \item Matern 3/2 kernel: $k^{m3/2}(x,y) = $ $\alpha(1+\sqrt{3}r) \exp(-\sqrt{3}r)$, where $r = \frac{\left \| x-y \right \|_2}{l}$ \cite{sriperumbudur2010hilbert}.
  \item Matern 5/2 kernel: $k^{m5/2}(x,y) = $ $\alpha(1+\sqrt{5}r+\frac{5}{3}r^2) \exp(-\sqrt{5}r)$, where $r = \frac{\left \| x-y \right \|_2}{l}$ \cite{sriperumbudur2010hilbert}.
  \item Linear combination of characteristic kernels by soft selection: $k^{lc}(x,y) = \sum_{i=1}^{K}\xi_i k_i(x,y)$, where $\xi_i \in [0, 1]$, $\sum_{i=1}^{K}\xi_i = 1$. $K=6$ and $k_i$ is one of the above kernel functions. 

\end{enumerate}

\subsection{Network Settings}
\label{sec:network-arc}
We use a fully-connected neural network on synthetic datasets 
and a ResNet on real image datasets. The architecture details 
are listed in Tables~\ref{tab:architecture_synthetic_datasets} --~\ref{ResNet_architectures_for_CelebA}. 

We adopt RMSprop optimizer~\cite{hinton2012lecture} with $\beta=0.99$ 
throughout all the experiments and a uniform distribution with 2 
dimensions for synthetic datasets and 128 dimensions for real image 
datasets as the input random noise. The learning rate $\eta$ is 0.0001 
for synthetic datasets, and 0.00005 for all other datasets. In terms of batch size for 
gradient descent, synthetic datasets are set to be the size of the whole 
dataset and the other datasets are set to be 64. For the kernel functions, we use the 
following default hyper-parameters: $\sigma = 10$ for Gaussian and 
Exponential kernels, $ \sigma= 100$ for Laplacian kernel, and 
$\alpha = 1$, $l = 10$ for Matern 3/2 and Matern 5/2 kernels.

For \mmdgan,  we used gradient penalty rather than weight clipping suggest by the author for the fair comparison, and set $\lambda$ to 10 that is the same to ours. The Adam optimizer with $\beta_1 = 0.5$, $\beta_2 = 0.9$ is applied to the \smmdgan, and with $\beta_1 = 0.5$, $\beta_2 = 0.999$ is applied to the \immdgan. \mmdgan and \veegan use the same optimizer as ours. For \mmdgan and \smmdgan, RBF mixture kernel with the parameter $\sigma_i \in \left \{ 1,2,4,8,16 \right \}$ was used. For \immdgan, RBF mixture kernel with the parameter  $\sigma_i \in \left \{ 1,\sqrt{2},2,2\sqrt{2},4 \right \}$ was used as suggested by the author.

\begin{table}[H]
\renewcommand\arraystretch{1.2}
    \caption{Architecture for synthetic datasets.}
    \label{tab:architecture_synthetic_datasets}
    \begin{minipage}[c]{0.5\linewidth}
        \centerline{
            \subtable{
                \setlength{\tabcolsep}{6mm}{
                    \begin{tabular}{c}
                    \toprule   
                    \midrule
                    $z\in \mathbb{R}^2\sim$ $\mathcal U[-1,1] $  \\
                    \midrule
                     Dense, 1024  \\\midrule
                     BN, ReLU \\\midrule
                     Dense, 1024 \\ \midrule 
                     BN, ReLU \\\midrule
                     Dense, 1024 \\\midrule 
                     BN, ReLU \\\midrule
                     Dense, 2\\\midrule  
                    \bottomrule
                    (a) generator
                    \end{tabular}
                }
                
            }
              
        }
    \end{minipage}%
    \hfill
    \begin{minipage}[c]{0.5\linewidth}
        \centerline{
            \subtable{
                \setlength{\tabcolsep}{6mm}{
                    \centering
                    \begin{tabular}{c}
                    \toprule   
                    \midrule
                    $x \in \mathbb{R}^2 $ \\
                    \midrule
                     Dense, 1024  \\\midrule
                     BN,  ReLU \\\midrule
                     Dense, 1024 \\ \midrule 
                     BN, ReLU \\\midrule
                     Dense, 2\\\midrule  
                    \bottomrule
                    (b) discriminator
                    \end{tabular}
                }
            }
        }
    \end{minipage}
    
\end{table}

\begin{figure}[H]
\centerline{\includegraphics[width=2.0cm]{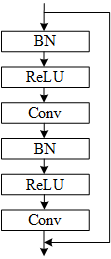}}
\caption{ResBlock architecture. We removed the BN layers in ResBlock for the discriminator, 
like \protect~\cite{miyato2018spectral}.} 
\label{tab:architecture_real_images}
\end{figure}

\begin{figure}[H]
    
    \begin{minipage}[t]{0.0\linewidth}

    \end{minipage}
    \hfill
    \begin{minipage}[t]{1.0\linewidth}
    
        \begin{table}[H]
        \renewcommand\arraystretch{1.2}
             \caption{ResNet architecture for the MNIST dataset. We use similar architectures to the one used in \protect~\cite{miyato2018spectral,gulrajani2017improved}.}
            \begin{minipage}[c]{0.45\linewidth}
                \centerline{
                    \subtable{
                        \setlength{\tabcolsep}{3mm}{
                        \begin{tabular}{c}
                              \toprule   
                                    \midrule
                                  $z \in \mathbb{R}^{128} \sim \mathcal U$[-1,1] \\ \midrule
                                    dense, 7 $\times$ 7 $\times$ 256   \\\midrule   
                                    ResBlock up 256  \\\midrule 
                                    ResBlock up 256  \\\midrule 
                                    BN, ReLU, 3$\times$3 conv, 3 Tanh \\\midrule
                                    \bottomrule
                                  (a) generator
                        \end{tabular}
                         
                        }
                     
                    }
                      
                }
            \end{minipage}%
    \hfill
    \begin{minipage}[c]{0.45\linewidth}
        \centerline{
            \subtable{
                \setlength{\tabcolsep}{3mm}{
                    \centering
                    \begin{tabular}{c}
                        \toprule 
                        \midrule
                        images $x \in \mathbb{R}^{28 \times 28 \times 1}$ \\ \midrule 
                        ResBlock down 128  \\ \midrule 
                        ResBlock down 128 \\ \midrule 
                        ResBlock 128 \\ \midrule 
                        ReLU \\ \midrule
                        Flatten\\ \midrule 
                        dense $\rightarrow \mathbb{R}^{128}$ \\ \midrule 
                        \bottomrule   
                        (b) discriminator
                        \end{tabular}
                }
            }
        }
    \end{minipage}
\end{table}
    
    \end{minipage}

    \label{zrotate}
\end{figure}

\begin{table}[H]
\renewcommand\arraystretch{1.2}
     \caption{ResNet architecture for the CIFAR-10 dataset.}
    \begin{minipage}[c]{0.5\linewidth}
        \centerline{
            \subtable{
                \setlength{\tabcolsep}{3mm}{
                    \begin{tabular}{c}
                    \toprule   
                    \midrule
                    $z \in \mathbb{R}^{128} \sim \mathcal U[-1,1]$ \\ \midrule   
                    dense, 4 $\times$ 4 $\times$ 256   \\\midrule   
                    ResBlock up 256  \\\midrule 
                    ResBlock up 256  \\\midrule 
                    ResBlock up 256  \\\midrule 
                    BN, ReLU, 3$\times$3 conv, 3 Tanh \\\midrule
                    \bottomrule 
                    (a) generator
                    \end{tabular}
                }
            }
              
        }
    \end{minipage}%
    \hfill
    \begin{minipage}[c]{0.5\linewidth}
        \centerline{
            \subtable{
                \setlength{\tabcolsep}{3mm}{
                    \centering
                    \begin{tabular}{c}
                    \toprule 
                    \midrule
                    RGB images $x \in \mathbb{R}^{32 \times 32 \times 3}$ \\ \midrule 
                    ResBlock down 128  \\ \midrule 
                    ResBlock down 128 \\ \midrule 
                    ResBlock 128 \\ \midrule 
                    ResBlock 128 \\ \midrule
                    ReLU \\ \midrule
                    Flatten\\ \midrule 
                    dense $\rightarrow \mathbb{R}^{128}$ \\ \midrule 
                    \bottomrule   
                    (b) discriminator
                    \end{tabular}
                }
            }
        }
    \end{minipage}
\end{table}

\begin{table}[H]
\renewcommand\arraystretch{1.2}
     \caption{ResNet architecture for the CelebA and the LSUN datasets.}\label{ResNet_architectures_for_CelebA}
    \begin{minipage}[c]{0.5\linewidth}
        \centerline{
            \subtable{
                \setlength{\tabcolsep}{3mm}{
                    \begin{tabular}{c}
                    \toprule   
                    \midrule
                    $z \in \mathbb{R}^{128} \sim \mathcal U[-1,1]$ \\ \midrule   
                    dense, 4 $\times$ 4 $\times$ 256   \\\midrule   
                    ResBlock up 256  \\\midrule 
                    ResBlock up 256  \\\midrule 
                    ResBlock up 256  \\\midrule
                    ResBlock up 256  \\\midrule 
                    BN, ReLU, 3$\times$3 conv, 3 Tanh \\\midrule
                    \bottomrule
                    (a) generator
                    \end{tabular}
                }
            }
              
        }
    \end{minipage}%
    \hfill
    \begin{minipage}[c]{0.5\linewidth}
        \centerline{
            \subtable{
                \setlength{\tabcolsep}{3mm}{
                    \centering
                    \begin{tabular}{c}
                    \toprule 
                    \midrule
                    RGB images $x \in \mathbb{R}^{64 \times 64 \times 3}$ \\ \midrule 
                    ResBlock down 128  \\ \midrule 
                    ResBlock down 128 \\ \midrule
                    ResBlock down 128 \\ \midrule 
                    ResBlock 128 \\ \midrule 
                    ResBlock 128 \\ \midrule
                    ReLU \\ \midrule
                    Flatten\\ \midrule 
                    dense $\rightarrow \mathbb{R}^{128}$ \\ \midrule 
                    \bottomrule   
                    (b) discriminator
                    \end{tabular}
                }
            }
        }
    \end{minipage}
\end{table}  

\subsection{Evaluation Metrics}
\label{sec:metrics}

\myparagraph{Synthetic Datasets}
To measure the degree of mode collapse, we follow the two metrics introduced 
in \cite{srivastava2017veegan}. The first metric is the \textit{number of 
modes captured} by the generator, which is counted if there is at least one 
generated sample within three standard deviations from the center of that mode. 
In our work, we count the number of modes captured using one standard deviation, 
which is different from \cite{srivastava2017veegan}. The reason is that we want 
to impose a stricter restriction to enhance the distinguish mode capture ability. The second metric is the \textit{number of 
high-quality samples}, which measures the proportion of generated samples within 
three standard deviations of the nearest mode. Similarly, for 2D SmileFace, we check whether the mode of the 
eye part of the face is captured. As discussed by 
\cite{lin2018pacgan}, all these metrics suffer from some 
disadvantages, i.e., they cannot evaluate how well the generated distribution is 
captured. To overcome this, we use a third-party trained classifier to calculate 
the \textit{KL} divergence between the real and the generated distributions. 

\myparagraph{Real Image Datasets}
We consider the following two metrics: 
\begin{itemize}
    
    \item The Inception Score (IS), proposed by \cite{salimans2016improved}, is an objective metric for evaluating the quality and diversity of generated samples. It utilizes a pre-trained Inception Net to calculate the $\text{exp}(\mathbb{E}_x[\mathbb{KL}(p(y\; | \; x)\; ||\; p(y))])$. It favors that the marginal distribution $p(y)$ has high entropy which means high diversity of generated samples, but the conditional label distribution $p(y|x)$ has low entropy (better quality). 

    \item The Fr\'{e}chet inception distance (FID), proposed by \cite{heusel2017gans}, 
    is a metric that computes the distance between the feature representation of 
    real data and generated samples by using a pre-trained inception model. A lower FID indicates higher similarity 
    between the real and the generated distributions. 
\end{itemize}

\subsection{Environment}
\label{sec:enviroment}
Experiments were run on a server with an Intel Xeon E5-2620 @2.10GHz 
CPU, TITAN X GPU, and 256GB RAM, running Ubuntu 18.04.3. 
Models were implemented in Python 3.6.9 and Tensorflow 2.2. 


\section{Additional Experiments}
\label{sec:additional-exp}

\subsection{More Synthetic datasets Experiments}
\label{sec:More_Synthetic_datasets_Experiments}
\begin{figure}[H] 
    \centering
    \includegraphics[scale=.9]{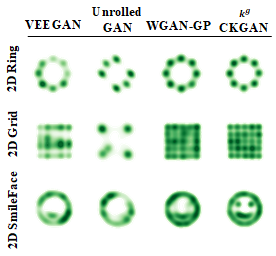} 
    \caption{Density plots of generated distributions by GAN variants using a relatively simple architecture.}
    \label{fig:simple_achitecture_synthetic_datasets} 
\end{figure}

Here we consider designing an artificial step that would be hard for \veegan but feasible for \ckgan. Note that in the following synthetic experiments, we equip \ckgan with Gaussian kernel $k^g$ ($\sigma=100$) and GP ($\lambda = 10$). \ckgan is supposed to satisfy the function set $\mathscr{F}$ is bounded in theory, but considering that GP may help mitigate the mode collapse, we also compare with \wgangp to exclude the influence of GP. The architectures for each dataset are shown in Tables~\ref{tab:architecture_ring_datasets}~and~\ref{tab:architecture_smile_datasets}. The visual results are shown in Figure~\ref{fig:simple_achitecture_synthetic_datasets}
and the quantities results are shown in Table~\ref{tab:table_simple_architecture_Synthetic}. We can observe that \ckgan is able to capture all modes and achieve a top two score in terms of HQ and KL. \unrolledgan on the 2D Ring has the best HQ score, since it drops all of generated samples into several modes.  \veegan and \wgangp can capture nearly all modes, but it fails to generate high quality samples.

\begin{table*}[!h]
  \centering
  \setlength{\tabcolsep}{3.0mm}{
  \caption{The number of modes (\#modes) captured (higher is better), the percentage of high-quality 
  (\%HQ) samples (higher is better) and KL divergence (lower is better) on synthetic datasets. Results are averaged over 10 trials.} 
  \label{tab:table_simple_architecture_Synthetic}
  \centering
  \resizebox{\linewidth}{!}{
  \begin{tabular}{lccccccccccccccc} 
    \toprule
    & \multicolumn{3}{c}{2D Ring} & \multicolumn{3}{c}{2D Grid} & \multicolumn{3}{c}{2D SmileFace} \\
    \cmidrule(r){2-4} \cmidrule(r){5-7} \cmidrule(r){8-11}
   \multirow{1}{*}{Methods}  & \#modes & \multirow{2}{*}{\%HQ} & \multirow{2}{*}{KL} & \#modes & \multirow{2}{*}{\%HQ} & \multirow{2}{*}{KL}  & \#modes & \multirow{2}{*}{\%HQ} & \multirow{2}{*}{KL}& \\
    & (max=8) & & & (max=25) & & & (max=2) &  &\\
    \midrule
    \unrolledgan &6.0&	\textbf{84.82}&	0.4941&	11.6&	75.57&	1.3147&	0&	0.0	&	0.2094\\
    \veegan &7.6&	63.50&	0.0253&	22.0&	20.12&	0.1430&	0.0&	1.43&	0.4546  \\
    \wgangp & \textbf{8.0}&	64.75&	\textbf{0.0014}&	24.6&	44.32&	0.0267&	0.6&	1.10&	0.0146 \\
    \ckgan $k^g$  &\textbf{8.0}&	71.88&	0.0016&	\textbf{25.0}&	\textbf{79.16}&	\textbf{0.0067}&	\textbf{2.0}&	\textbf{80.23}&	\textbf{0.0005}  \\
    
    \bottomrule
  \end{tabular}
  }
  }
\end{table*}


\begin{table}[H]
        \renewcommand\arraystretch{1.2}
             \caption{Architecture for 2D Ring and 2D Grid datasets. X is set to be 64 for 2D Ring and 1024 for 2D Grid}
              \label{tab:architecture_ring_datasets}
            \begin{minipage}[c]{0.5\linewidth}
                \centerline{
                    \subtable{
                        \setlength{\tabcolsep}{3mm}{
                        \begin{tabular}{c}
                              \toprule   
                                 \midrule
                                $z\in \mathbb{R}^2\sim$ $\mathcal U[-1,1] $  \\
                                \midrule
                                 Dense, X  \\\midrule
                                 BN, ReLU \\\midrule
                                 Dense, 2\\\midrule  
                                \bottomrule
                                (a) generator
                        \end{tabular}
                         
                        }
                     
                    }
                      
                }
            \end{minipage}%
    \hfill
    \begin{minipage}[c]{0.5\linewidth}
        \centerline{
            \subtable{
                \setlength{\tabcolsep}{3mm}{
                    \centering
                    \begin{tabular}{c}
                        \toprule 
                        \midrule
                    $x \in \mathbb{R}^2 $ \\
                    \midrule
                     Dense, X  \\\midrule
                     BN,  ReLU \\\midrule
                     Dense, 2\\\midrule  
                    \bottomrule
                    (b) discriminator
                        \end{tabular}
                }
           
            }
        }
    \end{minipage}
\end{table}

 \begin{table}[H]
        \renewcommand\arraystretch{1.2}
             \caption{Architecture for 2D smile dataset.}
             \label{tab:architecture_smile_datasets}
            \begin{minipage}[c]{0.45\linewidth}
                \centerline{
                    \subtable{
                        \setlength{\tabcolsep}{3mm}{
                        \begin{tabular}{c}
                              \toprule   
                                    \midrule
                                $z\in \mathbb{R}^2\sim$ $\mathcal U[-1,1] $  \\
                                \midrule
                                 Dense, 32  \\\midrule
                                 BN, ReLU \\\midrule
                                 Dense, 32  \\\midrule
                                 BN, ReLU \\\midrule
                                 Dense, 2\\\midrule  
                                \bottomrule
                                (a) generator
                        \end{tabular}
                         
                        }
                     
                    }
                      
                }
            \end{minipage}%
    \hfill
    \begin{minipage}[c]{0.45\linewidth}
        \centerline{
            \subtable{
                \setlength{\tabcolsep}{3mm}{
                    \centering
                    \begin{tabular}{c}
                        \toprule 
                        \midrule
                    $x \in \mathbb{R}^2 $ \\
                    \midrule
                     Dense, 32  \\\midrule
                     BN,  ReLU \\\midrule
                     Dense, 32  \\\midrule
                     BN,  ReLU \\\midrule
                     Dense, 2\\\midrule  
                    \bottomrule
                    (b) discriminator
                        \end{tabular}
                }
            }
        }
    \end{minipage}
\end{table}

\begin{table*}[!h]
  \small
  \centering
  \setlength{\tabcolsep}{5.6mm}{
  \caption{IS (higher is better) and FID (lower is better)  on real image datasets performed by \ckgan with SN. Results are averaged over 5 trials.} 
  \label{tab:table_small_images_sn_generation_metrics}
  \centering
  \resizebox{\linewidth}{!}{
    \begin{threeparttable}
      \begin{tabular}{lcccccccccc}
        \toprule

      \multirow{2}{*}{Methods} &\multicolumn{2}{c}{MNIST}&\multicolumn{2}{c}{CIFAR-10}&\multicolumn{2}{c}{CelebA}&\multicolumn{2}{c}{LSUN}             \\
        \cmidrule(r){2-3} \cmidrule(r){4-5} \cmidrule(r){6-7} \cmidrule(r){8-9} 
         & IS & FID & IS & FID & IS & FID & IS & FID  \\%
      \midrule
      Real data & 2.56 & 0.48 & 11.29 & 1.65 & 3.66 & 2.64 & 3.12 & 2.63\\
      \midrule

    \ckgan $k^g$	 &1.33 &	493.13  &  1.42 &	438.59 & 1.33 &  412.96 & 1.01 & 431.85\\
    \ckgan $k^l$ 	 &1.30 &	443.22 &  4.77 &	89.67 & 1.29 &	471.02 & 1.06 &	371.35\\
    \ckgan $k^{rbfm}$	 &1.30 &	495.70 & \textbf{5.24} &	87.90 & 1.73 &	355.43 & 1.27 &	\textbf{323.23}\\
    \ckgan $k^{e}$	 &2.37 &	9.68 & 1.44 &	433.12 & 1.04 &	373.97 & 1.54 &	464.15\\
    \ckgan $k^{m3/2}$	 &2.39 &  8.77 & 1.56 &	517.36 & \textbf{3.96} &	\textbf{186.26} & \textbf{1.99} & 415.27\\
    \ckgan $k^{m5/2}$	 &2.44 &	\textbf{7.59} & 1.09 &  382.83 & 1.44 &	432.22 & 1.09 &	418.17\\
    \ckgan $k^{lc}$	 &\textbf{2.47} &	8.08 & 4.71 &	\textbf{82.95} & 1.04 &	406.96 & 1.24 &	411.63\\
      \bottomrule

      \end{tabular}

  \end{threeparttable}
  }
  }
\end{table*}

\begin{table}[!h]
  \small
  \centering
  \caption{IS (higher is better) and FID (lower is better)  on higher resolution image datasets. Results are averaged over 5 trials.}
  \label{tab:table_big_images_generation_metrics}
  \setlength{\tabcolsep}{3.8mm}{
  \centering
  \begin{threeparttable}
    \begin{tabular}{lcccccccccc}
        \toprule
        \multirow{2}{*}{Methods} 
        & \multicolumn{2}{c}{CelebA} & \multicolumn{2}{c}{LSUN Bedroom} \\
        \cmidrule(r){2-3} \cmidrule(r){4-5}  
        & IS           & FID                   & IS    & FID \\
        \midrule
        Real data      & 4.33 & 1.38           & 2.62  & 0.24 \\
        \midrule
        VEEGAN         & 1.75 & 158.56         & - &- \\
        MMD GAN        & 2.25 & 53.97          & -  & 151.26 \\
        SMMD GAN       & - & 66.18  & -  & 85.14 \\
        IMMD GAN       & 2.58 & 21.5           & 2.12  & 31.43 \\
        CKGAN $k^{lc}$ & \textbf{3.13} & \textbf{19.95} & \textbf{2.46}  & \textbf{30.52} \\
        \bottomrule
    \end{tabular}
  \end{threeparttable}
  }
\end{table}

\begin{table}[!h]
  \small 
  \centering
  \caption{The performance of \ckgan using Gaussian kernel with different $\sigma$ values: 
  the number of modes (\#modes) captured (higher is better), the percentage of high-quality 
  (\%HQ) samples (higher is better), KL divergence (lower is better) on 2D Ring; IS (higher is better) and FID
  (lower is better) on MNIST.}
  \label{tab:table_parameter_comparision} \centering
  \setlength{\tabcolsep}{1.7mm}{
  \begin{tabular}{lccccc}
    \toprule
     \multirow{3}{*}{Parameters} & \multicolumn{3}{c}{2D Ring}&\multicolumn{2}{c}{MNIST}                  \\
    \cmidrule(r){2-4} \cmidrule(r){5-6} 
     & \#modes & \multirow{2}{*}{\%HQ} & \multirow{2}{*}{KL} & \multirow{2}{*}{IS}&\multirow{2}{*}{FID}\\%
    & (max=8) & \\
    \midrule
 Gaussian \, $\sigma=0.1   $  &     0.6	& 0.25	 & 4.5072	&1.29	 & 481.24 \\
 Gaussian \, $\sigma=1      $  &    1	& 2.10	& 8.0063	&1.30	 &477.62 \\
 Gaussian \, $\sigma=10    $  &   \textbf{8}	& \textbf{98.07}	& \textbf{0.0021}	                &\textbf{2.55}	 &\textbf{2.76} \\
 Gaussian \, $\sigma=100  $  &   6	& 82.22	& 0.0195	&2.41	 &6.69 \\
 Gaussian \, $\sigma=1000$  &   4.9	& 63.07	& 0.0558	&2.41	 &6.80 \\
    \bottomrule
  \end{tabular}
  }
\end{table}

\subsection{Additional Qualitative Results on Real Image Datasets}
\label{sec:more-qualitative}

\begin{figure}[H] 
    \centering
    \includegraphics[width=0.82\textwidth]{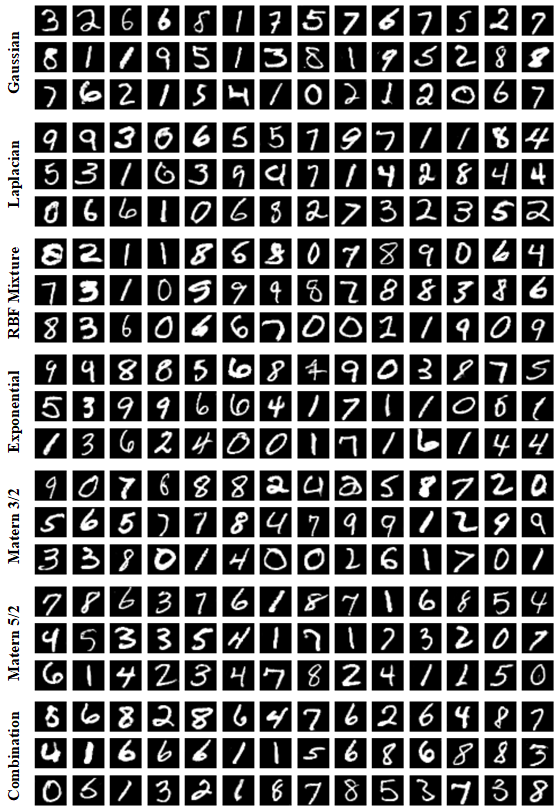} 
    \caption{Randomly generated samples by \ckgan on MNIST.}
    \label{fig:more_generatoin_mnist} 
\end{figure}

\begin{figure}[H] 
    \centering 
    \includegraphics[]{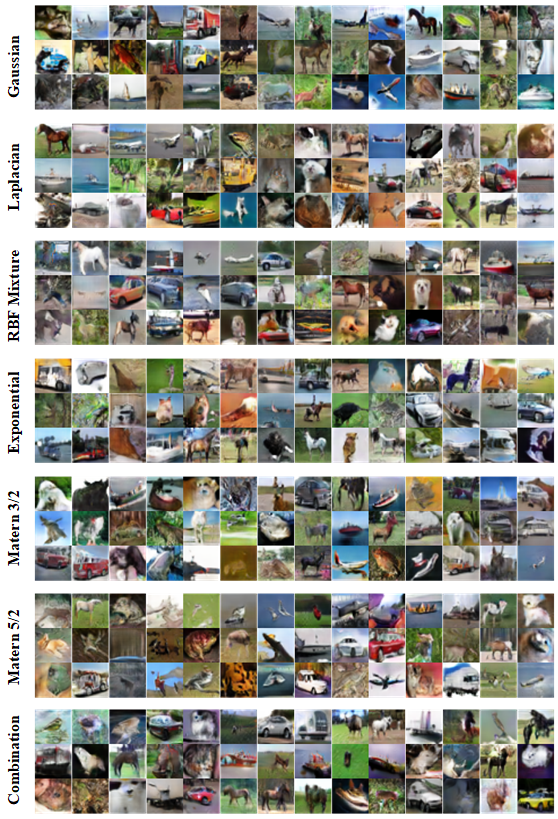} 
    \caption{Randomly generated samples by \ckgan on CIFAR-10.}    
    \label{fig:more_generatoin_cifar10} 
\end{figure}

\begin{figure}[H] 
    \centering 
    \includegraphics[]{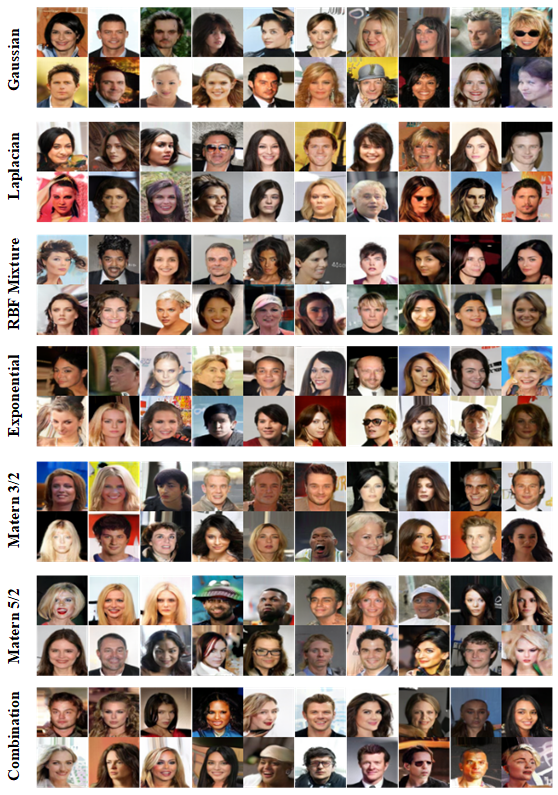} 
    \caption{Randomly generated samples by \ckgan on CelebA.}    
    \label{fig:more_generation_celeba} 
\end{figure}

\begin{figure}[H] 
    \centering 
    \includegraphics[]{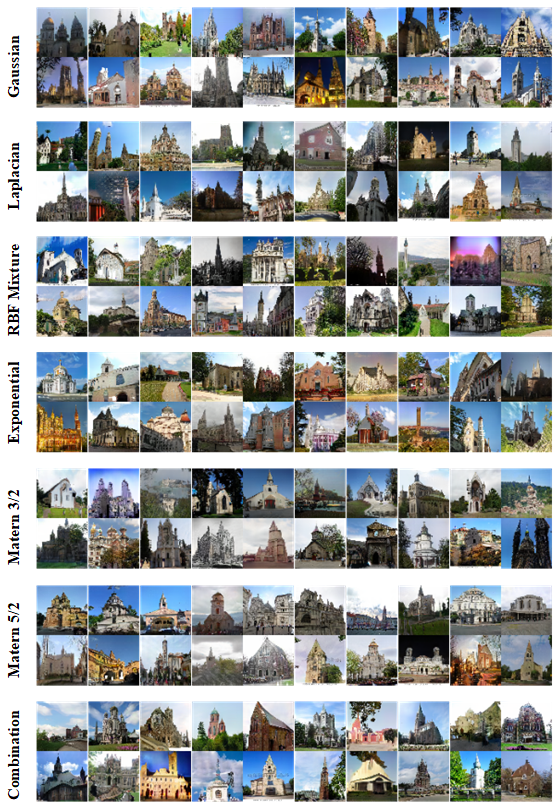} 
    \caption{Randomly generated samples by \ckgan on LSUN.}    
    \label{fig:more_generatoin_lsun} 
\end{figure}

\begin{figure}[H] 
    \centering 
    \includegraphics[width=0.9\textwidth]{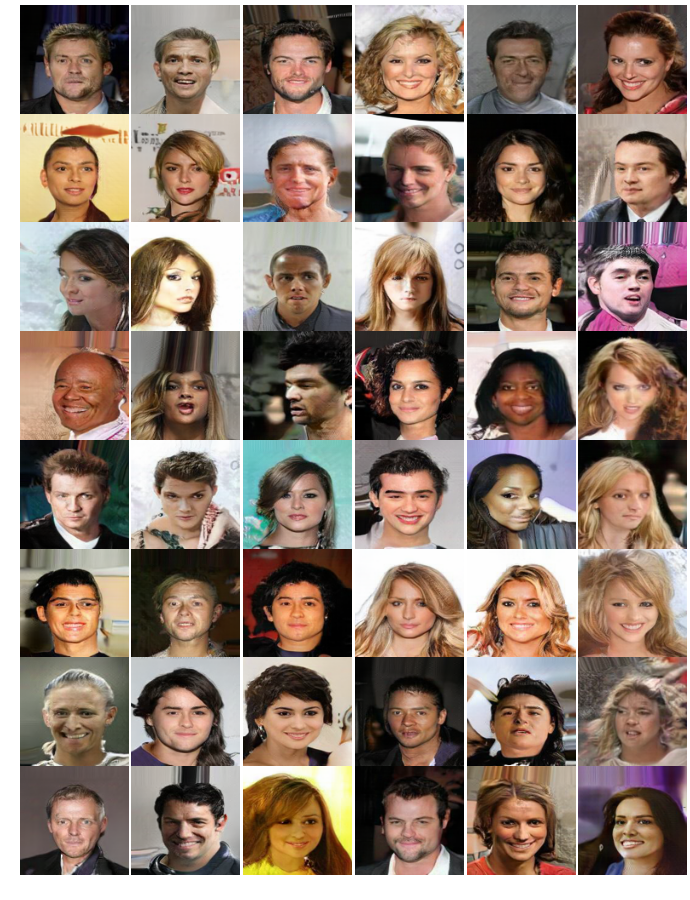} 
    \caption{Randomly generated samples by \ckgan $k^{lc}$ on 128 $\times$ 128 CelebA.}    
    \label{fig:more_generatoin_higher_celeba} 
\end{figure}

Figures~\ref{fig:more_generatoin_mnist} --~\ref{fig:more_generatoin_lsun} 
provide more generation samples of \ckgan, equipped with 
various kernels, on the four real image datasets. Figure~\ref{fig:more_generatoin_higher_celeba} shows the generated samples of \ckgan $k^{lc}$ on higher resolution image (128 $\times$ 128 CelebA) datasets. The size of dataset is reduced to 30,000. Table~\ref{tab:table_big_images_generation_metrics} shows the corresponding quantities metrics. the mark "-" on table means that the related method do not gain reasonable results and thus this metric are omitted. Due to the limitation of our hardware, we set batch size to 8 and add an upsampling layer to the generator used in 64 $\times$ 64 datasets.

\subsection{Effect of Kernel Parameters}
We evaluate the kernel parameters of \ckgan, and report the results of Gaussian kernel in Table~\ref{tab:table_parameter_comparision}, which shows the number of
modes captured, the percentage of high-quality, KL divergence on the 2D Ring dataset and 
IS and FID on the MNIST dataset. By varying the parameters, we observe that very small 
(e.g., $\sigma = 0.1$ or $1$ for Gaussian kernel) and very large values (e.g., $\sigma = 1000$ 
 for Gaussian kernel) do not perform well, while the values in the middle perform relatively better (e.g., $\sigma = 10$ or $100$). This indicates that we can obtain good results by 
equipping \ckgan with kernel parameters in the middle range. Based on the observed results 
in this set of experiments, we choose $\sigma = 10$ for Gaussian kernel.

\subsection{Generation Time}
Since we use the same network architecture for all the competitors, they report the 
same generation time. We test the generation time averaged over 1,000 images. The 
results are 13, 130, 153, and 153 microseconds on MNIST, CIFAR-10, CelebA, and LSUN, 
respectively.

\subsection{Applying Spectral Normalization to \ckgan}
Table~\ref{tab:table_small_images_sn_generation_metrics} shows the performance of \ckgan with SN (spectral normalization~\cite{miyato2018spectral}) on real image datasets.
We chose not to apply spectral normalization because the performance is mediocre (e.g., it reduces IS from 3.10 to1.04 and increases FID from 6.95 to 406.96 on CelebA). We can also observe that, on MNIST, some kernels with SN are unable to achieve reasonable results (higher FID) such as Gaussian kernel $k^g$ while linear combination of characteristic kernels $k^{lc}$ can gain relatively lower FID. This means that our kernel function selection method can automatically tune a kernel function for a specific situation.


\clearpage
\onecolumn

%% file: main.bbl
\begin{thebibliography}{10}

\bibitem{Arbel2018OnGR}
M.~Arbel, D.~Sutherland, M.~Binkowski, and A.~Gretton.
\newblock On gradient regularizers for mmd gans.
\newblock In {\em {NIPS}}, pages 6701--6711, 2018.

\bibitem{arjovsky2017wasserstein}
M.~Arjovsky, S.~Chintala, and L.~Bottou.
\newblock Wasserstein {GAN}.
\newblock {\em arXiv preprint arXiv:1701.07875}, 2017.

\bibitem{brock2018large}
A.~Brock, J.~Donahue, and K.~Simonyan.
\newblock Large scale {GAN} training for high fidelity natural image synthesis.
\newblock In {\em {ICLR}}, 2019.

\bibitem{chen2020reusing}
R.~Chen, W.~Huang, B.~Huang, F.~Sun, and B.~Fang.
\newblock Reusing discriminators for encoding: Towards unsupervised image-to-image translation.
\newblock In {\em {CVPR}}, pages 8168--8177, 2020.

\bibitem{fukumizu2008kernel}
K.~Fukumizu, A.~Gretton, X.~Sun, and B.~Sch{\"o}lkopf.
\newblock Kernel measures of conditional dependence.
\newblock In {\em {NIPS}}, pages 489--496, 2008.

\bibitem{goodfellow2014generative}
I.~Goodfellow, J.~Pouget-Abadie, M.~Mirza, B.~Xu, D.~Warde-Farley, S.~Ozair, A.~Courville, and Y.~Bengio.
\newblock Generative adversarial nets.
\newblock In {\em {NIPS}}, pages 2672--2680, 2014.

\bibitem{gretton2008kernel}
A.~Gretton, K.~Borgwardt, M.~J. Rasch, B.~Scholkopf, and A.~J. Smola.
\newblock A kernel method for the two-sample problem.
\newblock {\em arXiv preprint arXiv:0805.2368}, 2008.

\bibitem{gretton2012kernel}
A.~Gretton, K.~M. Borgwardt, M.~J. Rasch, B.~Sch{\"o}lkopf, and A.~Smola.
\newblock A kernel two-sample test.
\newblock {\em {JMLR}}, 13(1):723--773, 2012.

\bibitem{gulrajani2017improved}
I.~Gulrajani, F.~Ahmed, M.~Arjovsky, V.~Dumoulin, and A.~C. Courville.
\newblock Improved training of wasserstein {GAN}s.
\newblock In {\em {NIPS}}, pages 5767--5777, 2017.

\bibitem{heusel2017gans}
M.~Heusel, H.~Ramsauer, T.~Unterthiner, B.~Nessler, and S.~Hochreiter.
\newblock {GAN}s trained by a two time-scale update rule converge to a local nash equilibrium.
\newblock In {\em {NIPS}}, pages 6626--6637, 2017.

\bibitem{hinton2012lecture}
G.~Hinton.
\newblock Lecture 6d: a separate, adaptive learning rate for each connection. slides of lecture neural networks for machine learning.
\newblock Technical report, Slides of Lecture Neural Networks for Machine Learning, 2012.

\bibitem{ho2020denoising}
J.~Ho, A.~Jain, and P.~Abbeel.
\newblock Denoising diffusion probabilistic models.
\newblock {\em NeurIPS}, 33:6840--6851, 2020.

\bibitem{huang2024gan}
N.~Huang, A.~Gokaslan, V.~Kuleshov, and J.~Tompkin.
\newblock The {GAN} is dead; long live the {GAN}! a modern {GAN} baseline.
\newblock {\em NeurIPS}, 37:44177--44215, 2024.

\bibitem{karras2022elucidating}
T.~Karras, M.~Aittala, T.~Aila, and S.~Laine.
\newblock Elucidating the design space of diffusion-based generative models.
\newblock {\em NeurIPS}, 35:26565--26577, 2022.

\bibitem{krizhevsky2009learning}
A.~Krizhevsky, G.~Hinton, et~al.
\newblock Learning multiple layers of features from tiny images, 2009.

\bibitem{lecun1998gradient}
Y.~LeCun, L.~Bottou, Y.~Bengio, and P.~Haffner.
\newblock Gradient-based learning applied to document recognition.
\newblock {\em Proceedings of the IEEE}, 86(11):2278--2324, 1998.

\bibitem{li2017mmd}
C.-L. Li, W.-C. Chang, Y.~Cheng, Y.~Yang, and B.~P{\'o}czos.
\newblock {MMD GAN}: Towards deeper understanding of moment matching network.
\newblock In {\em {NIPS}}, pages 2203--2213, 2017.

\bibitem{li2019implicit}
C.-L. Li, W.-C. Chang, Y.~Mroueh, Y.~Yang, and B.~P{\'o}czos.
\newblock Implicit kernel learning.
\newblock {\em arXiv preprint arXiv:1902.10214}, 2019.

\bibitem{li2015generative}
Y.~Li, K.~Swersky, and R.~Zemel.
\newblock Generative moment matching networks.
\newblock In {\em {ICML}}, pages 1718--1727, 2015.

\bibitem{lin2018pacgan}
Z.~Lin, A.~Khetan, G.~Fanti, and S.~Oh.
\newblock Pac{GAN}: The power of two samples in generative adversarial networks.
\newblock In {\em {NIPS}}, pages 1498--1507, 2018.

\bibitem{liu2015deep}
Z.~Liu, P.~Luo, X.~Wang, and X.~Tang.
\newblock Deep learning face attributes in the wild.
\newblock In {\em {ICCV}}, pages 3730--3738, 2015.

\bibitem{metz2016unrolled}
L.~Metz, B.~Poole, D.~Pfau, and J.~Sohl-Dickstein.
\newblock Unrolled generative adversarial networks.
\newblock {\em arXiv preprint arXiv:1611.02163}, 2016.

\bibitem{miyato2018spectral}
T.~Miyato, T.~Kataoka, M.~Koyama, and Y.~Yoshida.
\newblock Spectral normalization for generative adversarial networks.
\newblock In {\em {ICLR}}, 2018.

\bibitem{10.2307/1428011}
A.~Müller.
\newblock Integral probability metrics and their generating classes of functions.
\newblock {\em {AAP}}, 29(2):429--443, 1997.

\bibitem{nowozin2016f}
S.~Nowozin, B.~Cseke, and R.~Tomioka.
\newblock f-{GAN}: Training generative neural samplers using variational divergence minimization.
\newblock In {\em {NIPS}}, pages 271--279, 2016.

\bibitem{salimans2016improved}
T.~Salimans, I.~Goodfellow, W.~Zaremba, V.~Cheung, A.~Radford, and X.~Chen.
\newblock Improved techniques for training {GAN}s.
\newblock In {\em {NIPS}}, pages 2234--2242, 2016.

\bibitem{scholkopf2002learning}
B.~Sch{\"o}lkopf, A.~J. Smola, et~al.
\newblock Learning with kernels: Support vector machines, regularization.
\newblock {\em Optimization, and Beyond. MIT press}, 1(2), 2002.

\bibitem{sriperumbudur2012empirical}
B.~K. Sriperumbudur, K.~Fukumizu, A.~Gretton, B.~Sch{\"o}lkopf, G.~R. Lanckriet, et~al.
\newblock On the empirical estimation of integral probability metrics.
\newblock {\em {EJS}}, 6:1550--1599, 2012.

\bibitem{sriperumbudur2010hilbert}
B.~K. Sriperumbudur, A.~Gretton, K.~Fukumizu, B.~Sch{\"o}lkopf, and G.~R. Lanckriet.
\newblock Hilbert space embeddings and metrics on probability measures.
\newblock {\em {JMLR}}, 11:1517--1561, 2010.

\bibitem{srivastava2017veegan}
A.~Srivastava, L.~Valkov, C.~Russell, M.~U. Gutmann, and C.~Sutton.
\newblock {VEEGAN}: Reducing mode collapse in {GAN}s using implicit variational learning.
\newblock In {\em {NIPS}}, pages 3308--3318, 2017.

\bibitem{su2019gan}
J.~Su.
\newblock {O-GAN}: Extremely concise approach for auto-encoding generative adversarial networks.
\newblock {\em arXiv preprint arXiv:1903.01931}, 2019.

\bibitem{wang2018improving}
W.~Wang, Y.~Sun, and S.~Halgamuge.
\newblock Improving {MMD-GAN} training with repulsive loss function.
\newblock In {\em {ICLR}}, 2019.

\bibitem{yu2015lsun}
F.~Yu, A.~Seff, Y.~Zhang, S.~Song, T.~Funkhouser, and J.~Xiao.
\newblock {LSUN}: Construction of a large-scale image dataset using deep learning with humans in the loop.
\newblock {\em arXiv preprint arXiv:1506.03365}, 2015.

\end{thebibliography}
